\pdfoutput=1
\documentclass[11pt]{article}

\usepackage[final]{acl}

\usepackage{times}
\usepackage{latexsym}
\definecolor{customgreen}{HTML}{0A9D51} % Define the custom green color
\usepackage{tcolorbox}
\usepackage[T1]{fontenc}
\usepackage[utf8]{inputenc}

\usepackage{microtype}

\usepackage{inconsolata}

\usepackage{graphicx}

\title{Beyond Prompting: An Efficient Embedding Framework \\ for Open-Domain Question Answering}

\author{Zhanghao Hu$^{1}$,
        Hanqi Yan$^{1}$,
        Qinglin Zhu$^{1}$\thanks{The corresponding author.},
        Zhenyi Shen$^{1}$,
        Yulan He$^{1,2}$,
        Lin Gui$^{1}$ \\
        $^1$King's College London \hspace{1cm}
        $^2$The Alan Turing Institute \\
        \texttt{\{zhanghao.hu, hanqi.yan, qinglin.1.zhu, zhenyi.shen\}@kcl.ac.uk} \\
        \texttt{\{yulan.he,lin.1.gui\}@kcl.ac.uk}
  }

\usepackage{xspace}
\usepackage{multicol}
\usepackage{amsmath}
\usepackage{comment}
\usepackage{amssymb}
\usepackage{booktabs}
\usepackage{multirow}
\usepackage{makecell}
\usepackage{graphicx}
\usepackage{listings}
\usepackage{enumitem}

\newcommand{\frameworkname}{EmbQA\xspace}
\newcommand{\embeddingname}{exploratory embedding\xspace}

\begin{document}
\maketitle
\begin{abstract}

Large language models (LLMs) have recently pushed open-domain question answering (ODQA) to new frontiers. However, prevailing retriever–reader pipelines often depend on multiple rounds of prompt-level instructions, leading to high computational overhead, instability, and suboptimal retrieval coverage. In this paper, we propose \frameworkname, an embedding-level framework that alleviates these shortcomings by enhancing both the retriever and the reader. Specifically, we refine query representations via lightweight linear layers under an unsupervised contrastive learning objective, thereby reordering retrieved passages to highlight those most likely to contain correct answers. Additionally, we introduce an exploratory embedding that broadens the model's latent semantic space to diversify candidate generation and employs an entropy-based selection mechanism to choose the most confident answer automatically. Extensive experiments across three open-source LLMs, three retrieval methods, and four ODQA benchmarks demonstrate that \frameworkname substantially outperforms recent baselines in both accuracy and efficiency.
\end{abstract}

\section{Introduction} 

Recent advances in large language models (LLMs) \cite{achiam2023gpt,dubey2024llama} have propelled Open-Domain Question Answering (ODQA) to new heights. A central strategy in ODQA involves retrieving relevant knowledge \cite{lei-etal-2023-unsupervised} and then integrating it with LLMs acting as readers to synthesize accurate answers. This retriever-reader approach has shown promise in overcoming the inherent limitations of LLMs \cite{mialon2023augmented}. 

Yet, prevailing retriever-reader architectures face two key limitations. 
First, retrievers \cite{karpukhin-etal-2020-dense,lei-etal-2023-unsupervised} yield abundant candidate passages, they fail to prioritize those containing definitive answers effectively. 
This is evidenced by their low ground truth recall in top-ranked results, where directly retraining retrievers or applying prompt-level reranking \cite{chuang-etal-2023-expand} proves impractical due to prohibitive computational costs \cite{zhuang-etal-2024-beyond} and inherited inefficiency from multi-turn processes. 
Second, while reader relies on multi-turn prompt-level strategies such as self-verification \cite{gao-etal-2023-enabling}, or additional summarization \cite{kim2024sure}, which requires expensively inference cost by LLM, leads to computational inefficiency and instability in the answer selection.

To address these limitations, we propose \textbf{Emb}edding-Driven Reranking and Answer Generation Framework for Open Domain \textbf{QA} Driven (\textbf{\frameworkname}), which utilises the embedding strategy to enhance both efficiency and accuracy in both the retriever and the reader.

In the retriever, we propose an embedding-level rerank framework that leverages candidate answers generated by LLMs to guide query refinement via unsupervised contrastive learning. Compared with existing LLM prompting-based reranking frameworks \cite{karpukhin-etal-2020-dense, lei-etal-2023-unsupervised}, which only focus on a few candidate passages due to the high computational cost of language inference, our proposed method enables the full exploration of the entire selected knowledge base through a learnable embedding layer. By mapping both queries and candidate sentences into the retrieval space and refining the query embedding with only a simple linear combination, our approach effectively reranks retrieved passages to prioritize those most likely to contain the correct answer.

In the reader, as many existing works suggest that there is a latent presence of prerequisite knowledge within the model’s parameter space \cite{ye2025limoreasoning}, and inserting a single compressed token can activate the neural pathways in LLM to generate the correct answer \cite{cheng2024xragextremecontextcompression}. Building on this, we propose an order-statistic-based measure for \embeddingname generation. This method allows us to explore statistically orthogonal directions by inserting only one token-sized embedding. It not only enhances diversity but also improves efficiency, as it eliminates the need for additional prompting rounds for summarisation or verification. By utilising perturbed predictive entropy, we can filter out uncertain answer candidates. 

\begin{figure*}[ht]
    \centering
    \resizebox{\linewidth}{!}{
    \includegraphics[width=\textwidth]{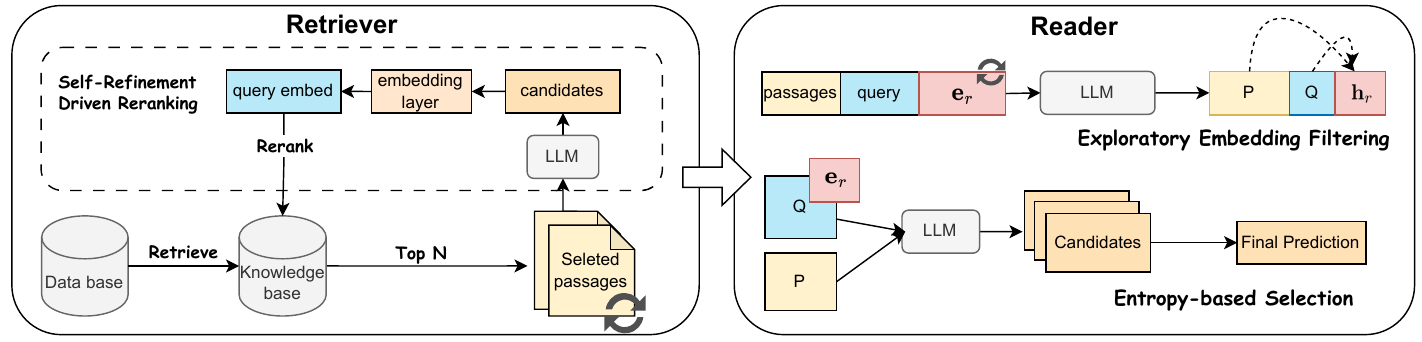}}
    \caption{Overview of the \frameworkname framework. \textit{Retriever} module constructs a knowledge base by retrieving passages from a large corpus and then refines the query via an embedding layer under unsupervised contrastive learning to prioritise passages rich in answer-critical cues. Then \textit{Reader} module integrates an exploratory embedding into the query to diversify candidate generation and employs an entropy-based selection mechanism to pick the final answer with the lowest uncertainty, ultimately enhancing both efficiency and overall performance in ODQA.}
    \label{fig:framework process}
\end{figure*}

In summary, our contributions are summarised as follows: 
\begin{enumerate} 
\item We develop an embedding-level rerank framework that leverages candidate answer guidance via unsupervised contrastive learning to optimise retrieval effectiveness. 
\item We propose an order-statistic-based single-token embedding strategy that activates latent knowledge within LLMs, reduces multi-turn prompting overhead and diversifies candidate generation.
\item Extensive experiments on three state-of-the-art open-sourced LLMs, and three retrieval methods across four ODQA datasets, demonstrate that our framework significantly outperforms existing prompt-level frameworks in both efficiency and accuracy. 
\end{enumerate}

\section{Related Work}

\noindent \textbf{Rerank of Retrieval-Augmented Generation in Open-Domain QA.} Open-domain question answering (ODQA) \cite{voorhees1999trec} typically adopts a retriever-reader framework \cite{chen-etal-2017-reading}, where a retriever selects relevant documents from extensive corpora, and a reader synthesises these into answers. Retrieval techniques generally fall into two categories: lexical methods (e.g., BM25 \cite{robertson2009probabilistic}) and dense models leveraging sentence embeddings (e.g., DPR \cite{karpukhin-etal-2020-dense} and Contriever \cite{lei-etal-2023-unsupervised}). In the era of large language models, Retrieval-Augmented Generation (RAG) similarly leverages external retrievers to augment language models by incorporating retrieved external knowledge, demonstrating effectiveness in reducing hallucinations and enhancing model accuracy \cite{JMLR:v24:23-0037,wang-etal-2024-searching}. However, both RAG-based methods and existing ODQA retrievers often face the challenge of underutilised retrieval capabilities, as top-ranked documents frequently lack comprehensive coverage of answers \cite{lazaridou2022internet,zhuang-etal-2024-beyond}. Reranking strategies aim to mitigate this issue by prioritising more relevant documents, thus enhancing answer coverage and accuracy \cite{zhuang2024setwise}. While recent approaches implement reranking at the prompt level \cite{10.1145/3626772.3657864,li-etal-2024-self-prompting}, these methods can be inefficient, time-consuming, and unstable \cite{wu2024passagespecific}. To address these limitations, we propose an embedding-level re-ranking mechanism that integrates seamlessly without requiring labelled data, providing a more efficient and stable alternative to prompt-level strategies.

\noindent \textbf{Prompt-level Framework of Reader in Open-Domain QA.} Prompt-level framework of Reader enhances language models by comparing multiple candidate solutions, either by selecting the best option \cite{kim2024sure} or synthesising several outputs for final prediction \cite{zhang-etal-2024-self-contrast} and self-verification at the end \cite{gao-etal-2023-enabling}. In ODQA, prevailing paradigms generate and evaluate candidate answers through summaries of retrieved content rather than directly querying large language models with original retrieved information \cite{giorgi-etal-2023-open, gao-etal-2023-enabling, kim2024sure}. However, these frameworks rely mainly on prompt-level mechanisms, which require multiple rounds of prompting, thus incurring inefficiencies and sensitivity to answer quality. Recent work has begun exploring a word-level framework via key term masking for self-correction \cite{wu-etal-2024-large}, yet this line of inquiry remains nascent. In contrast, our approach advances to the embedding level by incorporating diversity information during candidate generation to steer the model toward candidate sets more likely to contain the correct answer, and an entropy-based candidate filtering mechanism further ensures a more efficient selection process than prompt-level methods.

\section{Methodology}

\subsection{Overview and Problem Description}

\paragraph{Overview.} We propose \frameworkname, a two-stage framework that addresses the aforementioned limitations in open-domain QA. As illustrated in Figure~\ref{fig:framework process}, \frameworkname consists of: (1) \textit{Retriever} that refines the query through unsupervised contrastive learning, allowing it to effectively re-rank passages so that those potentially containing correct answers are prioritized; and (2) \textit{Reader} that broadens the semantic space for answer generation by injecting a lightweight \embeddingname derived from a normal distribution. This exploratory embedding nudges the model to discover a more diverse set of potential answers. Finally, we rely on an entropy-based criterion over the model’s output logits to select the best answer without resorting to multiple rounds of prompts. 

\paragraph{Problem Description.}
Open-domain question answering (ODQA) typically relies on external knowledge beyond a single context. Following the retriever–reader pipeline \cite{chen-etal-2017-reading,lee-etal-2019-latent}, we first form a \emph{knowledge base} \(\mathcal{B}\) by retrieving relevant passages from the large corpus \(\mathcal{D}\) given a query \(q\), then select the top-\(N\) passages \(\mathcal{C}_N\) from \(\mathcal{B}\) for candidate answer generation and subsequently refine the query representation $\mathbf{e}_{q_{\text{new}}}$ to re-rank \(\mathcal{B}\), emphasizing passages more likely to contain the correct answer. Then, we introduce a lightweight \emph{\embeddingname} \(\mathbf{e}_{r}\) to diversify the candidate answers. Finally, we compute logit-based entropy to select the best candidate as the final prediction.

\subsection{Retriever: Self-Refinement Driven Reranking}
\label{sec:Stage 1: Rerank and Self-Refine}
Despite significant progress, existing retrieval frameworks continue to struggle with prioritising passages containing definitive answers effectively. To address this, we propose a reranking framework that increases the likelihood of including ground-truth passages, thereby improving retrieval quality. Additionally, prior prompt-level reranking methods have proven inefficient and time-consuming \cite{zhuang2024setwise}. In contrast, \frameworkname adopts an embedding-based reranking strategy that is both more scalable and more efficient than previous prompt-level approaches \cite{li-etal-2024-self-prompting,zhuang-etal-2024-beyond}.

\textbf{Candidate Sentence Generation.} Given a query and its retrieved passages, we use a specialised prompt to generate $K$ candidate answers ${y} = [{y}_1, \dots, {y}_K]$ via an LLM. While prior studies \cite{lazaridou2022internet, weng-etal-2023-large} employ stochastic decoding to enhance diversity, we adopt an approach inspired by \cite{kim2024sure}, explicitly prompting the LLM to generate $K$ answer candidates. Following the empirical findings of \cite{kim2024sure}, which indicate no performance gains with larger $K$, we set $K=2$ in this research.

\textbf{Rerank by Unsupervised Contrastive Learning.}
Directly tuning a large retriever or relying on multiple rounds of prompt-based reranking can be computationally prohibitive. To address this, we propose a lightweight method that \emph{refines} the query representation, enabling more precise passage selection without retriever-wide fine-tuning.

Concretely, let a frozen retriever map the original query \(q\) and an LLM-generated candidate answer $y$ into a shared representation space, producing embeddings \(\mathbf{e}_q\) and \(\mathbf{e}_{y}\). We then form a new query embedding \(\mathbf{e}_{q_{\text{new}}}\) via a simple linear combination:
\begin{equation}
    \mathbf{e}_{q_{\text{new}}} = \mathbf{W}_1 \mathbf{e}_{y} + \mathbf{W}_2 \mathbf{e}_q,
\end{equation}

where \(W_1\) and \(W_2\) are the only trainable parameters. This design is far more efficient than modifying the entire retriever, while still capturing critical cues from the candidate sentence.

To learn \(\mathbf{W}_1\) and \(\mathbf{W}_2\), we adopt an unsupervised contrastive loss \cite{oord2018representation} that encourages \(\mathbf{e}_{q_{\text{new}}}\) to focus on passages containing the candidate answers from \({y}\). Specifically, we treat passages that contain at least one candidate as \emph{positive}, and those that do not as \emph{negative}. Because negatives are significantly more abundant, we sample them at a fixed ratio of 5:1 relative to positives in each training batch to maintain balance. Once the parameters are updated, we use \(\mathbf{e}_{q_{\text{new}}}\) to re-query the retriever, effectively \emph{re-ranking} the knowledge base \(\mathcal{B}\) so that passages with correct answers appear more prominently. This addresses the inefficiency of multi-turn prompt-level reranking and increases the likelihood that the subsequent Reader module receives high-quality evidence.

\subsection{Reader: Enhancing Generation via Exploratory Embedding}
\label{sec:Two-Tier Robustness Filtering}

Unlike existing ODQA patterns such as SuRe \cite{kim2024sure}, which relies on summarisation and prompt-level candidate selection strategies, \frameworkname removes summarisation and replaces prompt-level candidate selection with embedding-level entropy-based selection. Furthermore, our approach diversifies candidate generation via the \embeddingname mechanism with variance gate filtering, which guides the model to explore a broader semantic space.

\paragraph{Exploratory Embedding Filtering}
We introduce an \emph{\embeddingname} to diversify candidate generation. Suppose the LLM \(\mathcal{M}\) has an embedding dimension of \(D\). We sample a random vector \(\mathbf{e}_{r} \in \mathbb{R}^{D}\) from a standard normal distribution and concatenate it with the query \(q\) and retrieved context \(\mathcal{C}_N\) at inference time. We then extract the hidden representation \(\mathbf{h}_{r}\) corresponding to \(\mathbf{e}_{r}\) from the penultimate layer of \(\mathcal{M}\), following the practice of \citet{liu-etal-2024-fantastic} to capture sentence-level semantics.

Inspired by \citet{jain2014almost}, who shows that encouraging orthogonality among a set of vectors can be achieved by minimising their maximum inner products, we adopt the inner product as a measure of diversity. Our goal is to find an \(\mathbf{h}_r\) that is maximally \emph{inconsistent} (i.e., yields the smallest inner product) with the concatenated representation \(E(\mathcal{C}_N; q)\). However, directly optimising over the entire embedding space is computationally expensive—especially when considering tokens that have not yet been decoded. 
To address this, we assume that token embeddings follow a Gaussian distribution and derive an analytical approximation using order statistics. Specifically, we sort the elements of \(\mathbf{h}_r\) in descending order and define \(\Delta_i\) as the gap between the \(i\)-th and \((i+1)\)-th largest elements. We then show that minimising the expected inner product between \(\mathbf{h}_r\) and a set of Gaussian vectors is approximately equivalent to minimising the squared sum of the top-\(p\) differences:
$S_{\mathbf{e}_r} = \sum_{i=1}^{p} \Delta_{(i)}^2.$
(A detailed theoretical discussion justifying this approximation in relation to \citet{jain2014almost}'s claim is provided in Appendix~\ref{appendix:Theoretical Discussion on Embedding Space}.) We repeat the sampling process until we obtain an \(\mathbf{e}_r\) such that \(S_{\mathbf{e}_r}\) falls below a preset threshold \(T\).

\textbf{Entropy-Based Selection.}  
Once a suitable \(\mathbf{e}_r\) is obtained, we regenerate candidate answers \(\hat{y} = [\hat{y}_1, \dots, \hat{y}_K]\) using the LLM with the retrieved context \(\mathcal{C}_N\), the original query \(q\), and the selected exploratory embedding. Rather than relying on multi-turn prompts for candidate refinement, we leverage the LLM's own output uncertainties. Specifically, for each candidate answer, we compute a logit-based entropy score. Motivated by recent findings \cite{wu2024retrieval,wang-etal-2025-llms} that lower entropy correlates with higher confidence, we select the candidate with the lowest entropy:

\begin{equation}
\hat{a} \;=\; \operatorname*{argmin}_{\hat{y} \in \{\hat{y}_1, \dots, \hat{y}_K\}} \mathrm{Entropy}(\hat{y}).
\end{equation}

This embedding-level, entropy-based selection strategy eliminates the need for additional prompt rounds, making the answer generation process both efficient and robust.

\begin{table*}[!htbp]
    \centering
    \begin{tabular}{rcccccccccc}
        \toprule
        \multirow{2}{*}{\textbf{Method/Dataset}} & \multicolumn{2}{c}{\textbf{HotpotQA}} & \multicolumn{2}{c}{\textbf{2Wiki}} & \multicolumn{2}{c}{\textbf{NQ}} & \multicolumn{2}{c}{\textbf{WebQ}}  & \multicolumn{2}{c}{\textbf{Average}}\\
        \cmidrule(lr){2-3} \cmidrule(lr){4-5} \cmidrule(lr){6-7} \cmidrule(lr){8-9} \cmidrule(lr){10-11}
        & EM & F1 & EM & F1 & EM & F1 & EM & F1 & EM & F1\\
        \midrule
        No Retrieval & 20.8 & 29.1 & 12.2 & 16.2 & 20.6 & 26.6 & 17.2 & 25.8 & 17.7 & 24.4\\
        \midrule
        Retrieval Only & 25.4 & 37.2 & 16.6 & 21.1 & 26.0 & 32.8 & 22.2 & 31.2 & 22.6 & 30.6\\
        Chain-of-Thought & 27.0 & 39.8 & 15.4 & 21.8 & 27.2 & 33.5 & 28.8 & 37.8 & 24.6 & 33.2\\
        Self-Verification & 32.8 & 49.5 & 21.0 & 23.5 & 28.0 & 37.7 & 27.2 & 40.2 & 27.4 & 38.0\\
        SuRe & 38.8 & 53.5 & 23.8 & 31.0 & 36.6 & 47.9 & 34.4 & 48.5 & 33.4 & 45.3\\
        RPG & 37.9 & 49.2 & 24.6 & 33.8 & 36.6 & 50.5 & 34.2 & 47.3 & 33.3 & 45.2\\
        KnowTrace & 38.8 & 48.9 & 24.7 & 33.5 & 33.7 & 43.1 & 32.2 & 44.7 & 32.4 & 42.6\\
        \midrule
        \frameworkname (Ours) & \textbf{42.0} & \textbf{55.8} & \textbf{27.4} & \textbf{36.6} & \textbf{42.2} & \textbf{54.4} & \textbf{38.2} & \textbf{52.1} & \textbf{37.5} & \textbf{49.7}\\
        \bottomrule
    \end{tabular}
    \caption{Comparison of prompt-level frameworks on four open-domain QA datasets (HotpotQA, 2Wiki, NQ, and WebQ) using LLaMA 3.1. 
    % Exact Match (EM) and F1 scores are reported. 
    All methods retrieve the top-10 relevant passages using BM25. The \frameworkname framework outperforms existing prompt-level approaches across all datasets.}
    \label{tab:Comparison of prompt-level frameworks}
\end{table*}

\section{Experiments}

\subsection{Setups}
\label{sec:setups}

\textbf{Evaluation Datasets.} We evaluate \frameworkname on zero-shot QA across four ODQA datasets: HotpotQA~\cite{yang-etal-2018-hotpotqa}, 2WikiMulti-hopQA (2Wiki)~\cite{ho-etal-2020-constructing}, Natural Questions (NQ)~\cite{kwiatkowski-etal-2019-natural}, and  WebQuestions (WebQ)~\cite{berant-etal-2013-semantic}. For NQ and WebQ, we use their original test splits with the 21M English Wikipedia dump~\cite{karpukhin-etal-2020-dense} as the retrieval corpus. For all datasets, we adopt the implementation splits provided by~\citet{trivedi-etal-2023-interleaving} and~\citet{kim2024sure}, along with their respective document corpora.\\
\textbf{Metrics.} We use exact match (EM) and F1 score as evaluation metrics. Following~\citet{rajpurkar-etal-2016-squad}, we normalise predictions and gold answers by lowercasing and removing punctuation to ensure consistency.

\begin{table}[!ht]
    \centering
    \resizebox{\linewidth}{!}{%
    \setlength{\tabcolsep}{2pt}
    \begin{tabular}{c|l|cc|cc|cc|cc}
        \toprule
        \multirow{3}{*}{\rotatebox{90}{\textbf{Model}}} & \multirow{3}{*}{\makecell{\textbf{Retriever} \&\\ \textbf{Framework}}} & \multicolumn{8}{c}{\textbf{Dataset}} \\
        \cmidrule(lr){3-10}
         &  & \multicolumn{2}{c|}{HotpotQA} & \multicolumn{2}{c|}{2Wiki} & \multicolumn{2}{c|}{NQ} & \multicolumn{2}{c}{WebQ} \\
        \cmidrule(lr){3-4} \cmidrule(lr){5-6} \cmidrule(lr){7-8} \cmidrule(lr){9-10}
         &  & EM & F1 & EM & F1 & EM & F1 & EM & F1 \\
        \midrule
        \multirow{9}{*}{\rotatebox{90}{LLaMA3.1 8B-Ins}} 
        & BM25             & 25.4 & 37.2 & 16.6 & 21.1 & 26.0 & 32.8 & 22.2 & 31.2 \\
        & +SuRe            & 38.8 & 53.5 & 23.8 & 31.0 & 36.6 & 47.9 & 34.4 & 48.5 \\
        & +\frameworkname(ours)   & \textbf{42.0} & \textbf{55.8} & \textbf{27.4} & \textbf{36.6} & \textbf{42.2} & \textbf{54.4} & \textbf{38.2} & \textbf{52.1} \\
        \cmidrule(lr){2-10}
        & DPR              & 20.6 & 21.7 & 10.8 & 13.5 & 25.0 & 34.2 & 23.8 & 34.4 \\
        & +SuRe            & 25.0 & 31.9 & 14.2 & 16.0 & 38.8 & 52.3 & 36.0 & 49.6 \\
        & +\frameworkname(ours)   & \textbf{29.8} & \textbf{36.3} & \textbf{16.8} & \textbf{21.0} & \textbf{43.0} & \textbf{54.4} & \textbf{38.0} & \textbf{52.0} \\
        \cmidrule(lr){2-10}
        & Contriever       & 22.6 & 35.4 & 16.6 & 20.7 & 25.8 & 32.8 & 25.2 & 34.2 \\
        & +SuRe            & 33.8 & 50.6 & 21.0 & 29.3 & 39.0 & 52.8 & 34.4 & 48.5 \\
        & +\frameworkname(ours)   & \textbf{36.6} & \textbf{52.7} & \textbf{26.4} & \textbf{34.2} & \textbf{42.2} & \textbf{53.6} & \textbf{36.0} & \textbf{49.6} \\
        \midrule
        \multirow{9}{*}{\rotatebox{90}{Mistral v0.2 7B-Ins}} 
        & BM25             & 21.2 & 29.2 & 13.8 & 21.7 & 18.8 & 25.3 & 19.0 & 26.1 \\
        & +SuRe            & 32.2 & \textbf{46.1} & 17.8 & 30.1 & 35.2 & 45.1 & 31.6 & 45.7 \\
        & +\frameworkname(ours)   & \textbf{34.8} & 44.3 & \textbf{18.6} & \textbf{30.5} & \textbf{35.8} & \textbf{46.0} & \textbf{35.8} & \textbf{48.1} \\
        \cmidrule(lr){2-10}
        & DPR              & 7.8  & 11.0 & 3.8  & 4.5  & 22.2 & 26.7 & 18.8 & 27.7 \\
        & +Sure            & 15.0 & 21.8 & 6.4  & 8.5  & 40.0 & \textbf{51.8} & 32.6 & \textbf{47.7} \\
        & +\frameworkname(ours)   & \textbf{16.2} & \textbf{23.3} & \textbf{7.6}  & \textbf{9.6}  & \textbf{40.2} & 49.4 & \textbf{33.4} & 46.0 \\
        \cmidrule(lr){2-10}
        & Contriever       & 19.4 & 28.6 & 13.6 & 20.7 & 21.8 & 27.4 & 17.8 & 24.4 \\
        & +SuRe            & 28.0 & 41.6 & 17.2 & 25.4 & 39.8 & 51.6 & 30.2 & \textbf{45.0} \\
        & +\frameworkname(ours)   & \textbf{29.8} & \textbf{42.3} & \textbf{17.4} & \textbf{26.2} & \textbf{40.6} & \textbf{51.8} & \textbf{31.6} & 43.0 \\
        \midrule
        \multirow{9}{*}{\rotatebox{90}{Qwen 2.5 7B-Ins}} 
        & BM25             & 28.6 & 37.1 & 20.2 & 24.1 & 24.0 & 29.4 & 22.6 & 31.4 \\
        & +Sure            & 43.6 & 54.7 & 28.4 & \textbf{34.1} & 41.6 & 49.0 & 36.6 & 47.3 \\
        & +\frameworkname(ours)   & \textbf{44.6} & \textbf{55.6} & \textbf{28.8} & 33.8 & \textbf{42.4} & \textbf{49.2} & \textbf{38.2} & \textbf{48.7} \\
        \cmidrule(lr){2-10}
        & DPR              & 8.8  & 9.8  & 5.6  & 7.1  & 29.2 & 32.6 & 25.6 & 31.1 \\
        & +Sure            & 21.8 & 27.3 & 12.2 & 16.1 & 45.4 & 54.6 & 38.4 & 49.6 \\
        & +\frameworkname(ours)   & \textbf{22.6} & \textbf{29.1} & \textbf{13.8} & \textbf{17.3} & \textbf{45.8} & \textbf{54.7} & \textbf{38.6} & \textbf{50.1} \\
        \cmidrule(lr){2-10}
        & Contriever       & 27.0 & 34.0 & 17.6 & 20.0 & 26.6 & 31.9 & 21.0 & 29.1 \\
        & +Sure            & 38.8 & \textbf{50.3} & 23.8 & 30.4 & 44.0 & \textbf{52.9} & 36.4 & 48.1 \\
        & +\frameworkname(ours)   & \textbf{39.0} & 50.2 & \textbf{24.4} & \textbf{30.9} & \textbf{45.2} & 50.5 & \textbf{37.0} & \textbf{48.6} \\
        \bottomrule
    \end{tabular}%
    }
    \caption{Exact Match (EM \%) and F1 score performance of LLaMA 3.1, Mistral v0.2, and Qwen 2.5 across HotpotQA, 2Wiki, NQ, and WebQ datasets. Each model is evaluated using three retrieval methods: BM25 (lexical retriever), DPR, and Contriever (dense retrievers). Results are reported for retrieval-only, SuRe, and our proposed framework. Across all models, retrievers, and datasets, our framework consistently outperforms both the SuRe baseline and retrieval-only approaches.}
    \label{tab:Exact Match (EM) and F1 score performance of LLaMA 3.1, Mistral v0.2, and Qwen 2.5}
\end{table}

\noindent \textbf{Baselines.} We compare \frameworkname with the following methods: (1) \textit{No Retrieval} generates answers applying an LLM in a closed-book setting without retrieved passages. (2) \textit{Retrieval Only} appends retrieved passages to the input prompt. (3) \textit{Chain-of-Thoughts} \cite{NEURIPS2022_8bb0d291,NEURIPS2022_9d560961} augments the prompt with zero-shot chain-of-thought reasoning. (4) \textit{Self-Verification} \cite{weng-etal-2023-large} generates multiple answer candidates via random sampling and selects the most plausible one by verifying its reasoning using conditional masking. (5) \textit{SuRe} \cite{kim2024sure} produces candidate answers and selects the most plausible one by conditional summarisation as the final prediction. (6) \textit{RPG} \cite{lyu-etal-2024-retrieve} iteratively guides large language models to generate more relevant and factually grounded outputs by planning with selected evidence. (7) \textit{KnowTrace} \cite{li2025knowtrace} enables large language models to construct question-specific knowledge graphs through iterative knowledge tracing progressively.

\noindent \textbf{Implementation Details.} Our framework requires modifications at the embedding level, which necessitates the use of open-sourced LLMs. We conduct experiments with three state-of-the-art models: 
LLaMA-3.1-8B-Instruct \cite{dubey2024llama}, Mistral-7B-Instruct-v0.2 \cite{jiang2023mistral}, and Qwen2.5-7B-Instruct \cite{yang2024qwen2}. We set the decoding temperature to 0.0 to ensure greedy decoding, to eliminate the effect of random sampling \cite{sun-etal-2023-chatgpt}. For retrieval, we employ three approaches: a lexical-based retriever (BM25) \cite{robertson2009probabilistic} and two dense retrievers (DPR-multi \cite{karpukhin-etal-2020-dense} and Contriever \cite{lei-etal-2023-unsupervised}). We use Elasticsearch for BM25 and the BEIR toolkit for DPR and Contriever,\footnote{\url{https://www.elastic.co/}, \url{https://github.com/beir-cellar/beir}} respectively. In our framework, an initial retriever retrieves candidate passages, which are then reranked based on modifications at the embedding level. Notably, when BM25 is used as the initial retriever—owing to its lexical nature and inability to generate sentence embeddings—we employ Contriever for reranking; in contrast, when DPR or Contriever serves as the initial retriever, the same model is used throughout. We use consistent prompts across all datasets (Appendix \ref{sec: appendix prompt design}) and fix $K=2$ in all experiments following \citet{kim2024sure}. Although iterative reranking is theoretically possible, we perform only a single reranking pass given the limited performance gains relative to the increased computational cost. In the exploratory embedding stage, a variance gating threshold of $0.05$ is applied.

\subsection{Main Results}

\paragraph{\frameworkname Outperforms Prompt-Level Methods.} Table \ref{tab:Comparison of prompt-level frameworks} presents the performance of lines of prompt-level frameworks on four open-domain QA datasets using LLaMA 3.1 with BM25-based retrieval of the top-10 passages. Notably, augmenting retrieved passages with prompting generally improves ODQA accuracy over a pure retrieval strategy. However, our proposed embedding-level framework consistently outperforms these prompt-level approaches across all datasets. For instance, on HotpotQA, our method achieves an Exact Match (EM) of 42.0 and an F1 score of 55.81, representing improvements of approximately 3.2 and 2.3 points over the best prompt-level baseline (SuRe). Similar gains are observed on the remaining datasets, underscoring the efficacy of leveraging embedding-level information to enhance LLM performance in open-domain QA tasks.

\paragraph{Robust Generality of \frameworkname Across Setups.} Table~\ref{tab:Exact Match (EM) and F1 score performance of LLaMA 3.1, Mistral v0.2, and Qwen 2.5} further demonstrates the broad compatibility of our embedding-level framework across three LLMs, four QA datasets, and three retrieval methods. \frameworkname consistently outperforms both retrieval-only and SuRe \cite{kim2024sure} baselines in nearly all settings. For example, on LLaMA 3.1 with DPR, \frameworkname achieves 29.8 EM and 36.34 F1 on HotpotQA, significantly outperforming BM25+SuRe (4.8 EM and 4.46 F1). While performance gains on Mistral and Qwen are relatively modest, this is primarily due to their overconfidence in the evaluated datasets, which distorts the token embedding distribution and complicates threshold selection for final answer prediction. We note that refining this threshold could further enhance performance\footnote{Further analysis and details are in Appendix {\ref{appendix:Overconfidence Analysis of Mistral and Qwen}}}. These findings underscore the robustness and generalizability of our approach in open-domain QA.

\paragraph{Efficiency.} 
Table~\ref{tab:Comparison of execution time and token throughput} compares execution time and output token usage between SuRe~\cite{kim2024sure} and our framework across four datasets using LLaMA 3.1 with Contriever. Our method consistently reduces both query time and token cost. For instance, on HotpotQA, EmbQA completes a query in $0.53$ minutes using $0.99k$ output tokens, compared to $1.56$ minutes and $3.51k$ tokens for SuRe \cite{kim2024sure}. This efficiency stems from fundamental differences in multi-turn prompting design. SuRe requires 7 LLM prompts per question—three stages (summary generation, validity checking, ranking), each applied individually to two candidates. In contrast, EmbQA achieves comparable answer diversity and quality using only 2 prompts: one for initial generation and one for enhanced exploration. Contrastive re-ranking and entropy-based selection are performed without additional LLM queries. As a result, EmbQA reduces total LLM calls by 66\% and per-step cost by roughly 50\%, yielding an estimated 33\% overall runtime compared to SuRe\footnote{Detailed comparison and motivations are in Appendix \ref{appendix:Detailed Comparison of Multi-turn Prompting Costs}}. Similar efficiency gains are observed across 2Wiki, NQ, and WebQ datasets, underscoring the superior computational efficiency of our approach.

\begin{table}[ht]
    \centering
    \resizebox{\linewidth}{!}{
    \setlength{\tabcolsep}{2pt}
    \begin{tabular}{llcc}
        \toprule
        \textbf{Dataset} & \textbf{Method} & \makecell{\textbf{Time}/\textbf{query (min) ↓}} & \makecell{\textbf{\textbf{Tokens}}  \textbf{/query ↓}} \\
        \midrule
        \multirow{2}{*}{HotpotQA}      & SuRe  & 1.56  & $3.51k$ \\
                                   & \frameworkname(ours)  & \textbf{0.53}  &  $\textbf{0.99k}$\\
        \midrule
        \multirow{2}{*}{2Wiki}    & SuRe  & 1.57  & $3.43k$ \\
                                   & \frameworkname(ours)  & \textbf{0.54}  &  $\textbf{1.20k}$\\
        \midrule
        \multirow{2}{*}{NQ}    & SuRe  & 1.43  &   $4.39k$ \\
                                   & \frameworkname(ours)  &\textbf{ 0.54}  & $\textbf{0.84k}$ \\
        \midrule
        \multirow{2}{*}{WebQ} & SuRe  & 1.58  & $3.91k$ \\
                                   & \frameworkname(ours)  &\textbf{ 0.56}  & $\textbf{1.31k}$ \\
        \bottomrule
    \end{tabular}}
    \caption{Comparison of execution time and output token requirement per query between SuRe and our proposed framework \frameworkname across four open-domain QA datasets (HotpotQA, 2Wiki, NQ, and WebQ), using LLaMA 3.1 with Contriever retrieval. Our method significantly reduces query time and output token requirement in all datasets.}
    \label{tab:Comparison of execution time and token throughput}
\end{table}

\section{Analysis}

\paragraph{Ablation Studies.}
\begin{figure*}[ht]
    \centering
    \includegraphics[width=\textwidth]{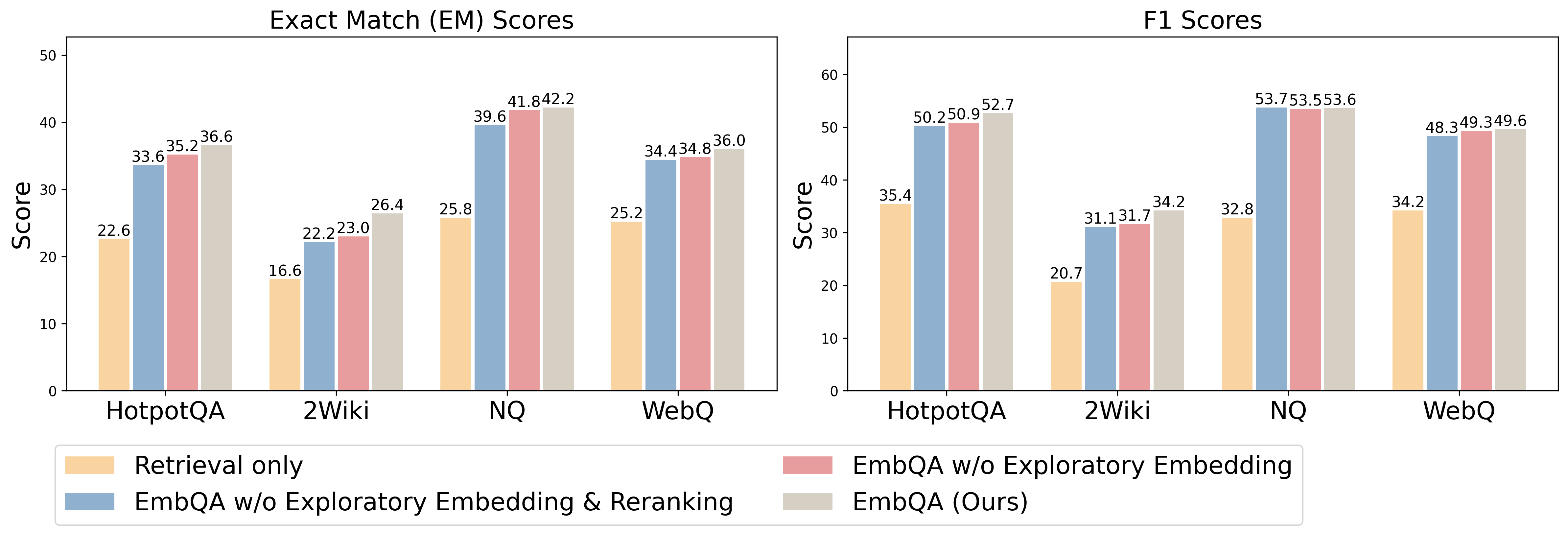}
    \caption{Ablation study on HotpotQA, 2Wiki, NQ, and WebQ datasets using LLaMA 3.1 with Contriever retrieval. The results compare four settings: (1) Retrieval Only, (2) Full \frameworkname, (3) \frameworkname without the \embeddingname module, and (4) \frameworkname without both the \embeddingname and reranking modules. Each component contributes crucially to the overall performance, as evidenced by incremental improvements in Exact Match (EM) and F1 scores.}
    \label{fig:fusion_study}
\end{figure*}

We conducted an ablation study to assess the contribution of each component in our overall framework. We evaluated four configurations: (1) Retrieval Only, (2) \frameworkname (our full framework), (3) \frameworkname ~{\bf w/o} \embeddingname (without the exploratory embedding module), and (4) \frameworkname ~{\bf w/o} exploratory embedding \& rerank (without both the \embeddingname and the reranking module)\footnote{Additional analysis on contrastive learning and the embedding refinement are in Appendix \ref{appendix:Further Fusion Study on Contrastive Learning and the Embedding Refinement}.}. Figure~\ref{fig:fusion_study} reports the Exact Match (EM) and F1 scores on HotpotQA, 2Wiki, NQ, and WebQ. The results indicate that each component contributes additively to performance improvements. For example, on HotpotQA, the full \frameworkname model achieves 36.6 EM and 52.7 F1, which is substantially higher than the 22.6 EM and 35.4 F1 obtained in the Retrieval Only setting. Removing the \embeddingname module results in a performance drop to 35.2 EM and 50.9 F1, and further removing the reranking module degrades the scores to 33.6 EM and 50.2 F1. Similar trends across all datasets demonstrate a sequential degradation in performance as each module is removed, highlighting the additive contributions of each component in our framework.

\begin{table*}[ht]
    \centering
    \resizebox{\textwidth}{!}{
    \begin{tabular}{lcc|cc|cc|cc}
        \toprule
        \multirow{2}{*}{\makecell{\textbf{Retriever\&} \\ \textbf{Rerank Framework}}} & \multicolumn{2}{c|}{\textbf{HotpotQA}} & \multicolumn{2}{c|}{\textbf{2Wiki}} & \multicolumn{2}{c|}{\textbf{NQ}} & \multicolumn{2}{c}{\textbf{WebQ}} \\
        \cmidrule(lr){2-3} \cmidrule(lr){4-5} \cmidrule(lr){6-7} \cmidrule(lr){8-9}
         & \makecell{\textbf{Avg. GT} \\ \textbf{@Top-10}} & \makecell{\textbf{Time} \\ \textbf{/Query(s)}} 
         & \makecell{\textbf{Avg. GT} \\ \textbf{@Top-10}} & \makecell{\textbf{Time} \\ \textbf{/Query(s)}}
         & \makecell{\textbf{Avg. GT} \\ \textbf{@Top-10}} & \makecell{\textbf{Time} \\ \textbf{/Query(s)}}
         & \makecell{\textbf{Avg. GT} \\ \textbf{@Top-10}} & \makecell{\textbf{Time} \\ \textbf{/Query(s)}} \\
        \midrule
        BM25 & 1.16 & --   & 0.81 & --   & 1.50 & --   & 1.88 & -- \\
        +Prompt Level & 1.06 & 12.52 & 1.09 & 12.62 & 1.62 & 12.65 & 2.70 & 12.69 \\
        +Embedding Level (Ours) & \textbf{1.42} & \textbf{1.33}  & \textbf{1.21} & \textbf{1.54}  & \textbf{2.57} & \textbf{1.90}  & \textbf{4.18} & \textbf{2.31} \\
        \midrule
        DPR & 0.28 & --   & 0.30 & --   & 1.79 & --   & 3.04 & -- \\
        +Prompt Level & 0.64 & 13.23 & 0.34 & 12.52 & 1.75 & 12.66 & 3.68 & 12.63 \\
        +Embedding Level (Ours) & \textbf{1.01} & \textbf{2.42}  & \textbf{1.13} & \textbf{1.27}  & \textbf{2.41} & \textbf{2.00}  & \textbf{4.25} & \textbf{2.22} \\
        \midrule
        Contriever & 1.47 & --   & 0.99 & --   & 1.98 & --   & 2.87 & -- \\
        +Prompt Level & 1.37 & 12.54 & 1.36 & 12.95 & 2.01 & 13.16 & 3.02 & 13.05 \\
        +Embedding Level (Ours) & \textbf{1.87} & \textbf{1.12}  & \textbf{1.49} & \textbf{2.93}  & \textbf{2.55} & \textbf{2.12}  & \textbf{4.31} & \textbf{2.69} \\
        \bottomrule
    \end{tabular}}
    \caption{Retrieval analysis on HotpotQA, 2Wiki, NQ, and WebQ datasets using LLaMA3.1. For each dataset, the two metrics are: \textbf{Average Ground Truth Passages in Top-10} (the average number of ground-truth passages present among the top-10 retrieved results) and \textbf{Time Consumption Per Query (seconds)} (the time taken for processing each query in seconds). Higher values in the first metric indicate that our reranking framework surfaces relevant passages more effectively compared to the BM25 baseline.}
    \label{tab:retrieval_analysis}
\end{table*} 

\paragraph{Why Prompt-Level Rerank Framework Fail in Existing ODQA Framework?} Existing ODQA systems \cite{kim2024sure} have demonstrated that prompt-level reranking can be ineffective or even detrimental to performance. We posit that this failure stems from the inability of prompt-level reranking to reliably elevate ground-truth passages among the top-10 retrieved results, and in some cases, it may even reduce their presence. Table~\ref{tab:retrieval_analysis} presents a comparison between our embedding-level rerank framework and a state-of-the-art prompt-level approach \cite{zhuang-etal-2024-beyond} across four datasets (HotpotQA, 2Wiki, NQ, and WebQ) using three different retrievers (BM25, DPR, and Contriever). Two key metrics are reported: (i) the average number of ground-truth passages in the top-10 results, and (ii) the per-query processing time (in seconds). Existing prompt-level rerank frameworks often struggle to effectively enhance these metrics, and in some cases, they even weaken them, resulting in unstable downstream ODQA performance. For example, on the HotpotQA dataset with BM25, the prompt-level method reduces the average ground-truth count from 1.16 to 1.06, potentially impairing downstream ODQA performance. In contrast, our embedding-level approach increases this count from 1.16 to 1.42. Similar trends are observed with DPR and Contriever retrievers, underscoring the broad applicability of our approach. Moreover, our experiments reveal that when substituting our embedding-level framework with the prompt-level alternative \cite{zhuang-etal-2024-beyond} within the overall system, performance degrades significantly across all datasets and retrievers (Table~\ref{tab:retrieval_rerank_comparison}). In terms of efficiency, our embedding-level framework dramatically reduces query processing time. For instance, with BM25 on HotpotQA, the prompt-level rerank requires 12.52 seconds per query, whereas our method reduces this to 1.33 seconds. To further assess generalisation to closed-source LLMs, we additionally apply our reranking module to the GPT-4o-mini API model without modifying internal model parameters. As shown in Appendix~\ref{appendix:closedsource_rerank}, integrating our reranker with the SuRe framework \cite{kim2024sure} consistently improves EM/F1 across four ODQA benchmarks, confirming its applicability to API-based systems as well.

\begin{table}[ht]
\centering
\resizebox{\linewidth}{!}{
\begin{tabular}{lcc|cc|cc|cc}
\toprule
\multirow{2}{*}{\makecell{\textbf{Retriever} \\ \textbf{\& Rerank Module}}} 
& \multicolumn{2}{c|}{\textbf{HotpotQA}} 
& \multicolumn{2}{c|}{\textbf{2Wiki}} 
& \multicolumn{2}{c|}{\textbf{NQ}} 
& \multicolumn{2}{c}{\textbf{WebQ}} \\
\cmidrule(lr){2-3} \cmidrule(lr){4-5} \cmidrule(lr){6-7} \cmidrule(lr){8-9}
& EM & F1 & EM & F1 & EM & F1 & EM & F1 \\
\midrule
BM25 & 25.4 & 37.2 & 16.6 & 21.1 & 26.0 & 32.8 & 22.2 & 31.2 \\
+ Prompt-Level & 39.6 & 52.4 & 19.8 & 28.7 & 39.8 & 51.8 & 36.6 & 50.1 \\
+ \textbf{Embedding-Level} & \textbf{42.0} & \textbf{55.8} & \textbf{27.4} & \textbf{36.6} & \textbf{42.2} & \textbf{54.4} & \textbf{38.2} & \textbf{52.1} \\
\midrule
DPR & 20.6 & 21.7 & 10.8 & 13.5 & 25.0 & 34.2 & 23.8 & 34.4 \\
+ Prompt-Level & 24.6 & 30.0 & 11.6 & 18.8 & 39.8 & 52.3 & 37.6 & 48.2 \\
+ \textbf{Embedding-Level} & \textbf{29.8} & \textbf{36.3} & \textbf{16.8} & \textbf{21.0} & \textbf{43.0} & \textbf{54.4} & \textbf{38.0} & \textbf{51.9} \\
\midrule
Contriever & 22.6 & 35.4 & 16.6 & 20.7 & 25.8 & 32.8 & 25.2 & 34.2 \\
+ Prompt-Level & 30.2 & 47.3 & 17.6 & 25.9 & 39.8 & 52.2 & 34.6 & 48.7 \\
+ \textbf{Embedding-Level} & \textbf{36.6} & \textbf{52.7} & \textbf{26.4} & \textbf{34.2} & \textbf{42.2} & \textbf{53.6} & \textbf{36.0} & \textbf{49.6} \\
\bottomrule
\end{tabular}
}
\caption{QA performance (EM/F1) when integrating prompt-level and embedding-level EmbQA rerank frameworks with BM25, DPR, and Contriever retrievers on four ODQA datasets using LLaMA 3.1.}
\label{tab:retrieval_rerank_comparison}
\end{table}

\paragraph{Effect of Exploration with Exploratory Embedding.}

\begin{figure}[ht]
    \centering
    \includegraphics[width=\linewidth]{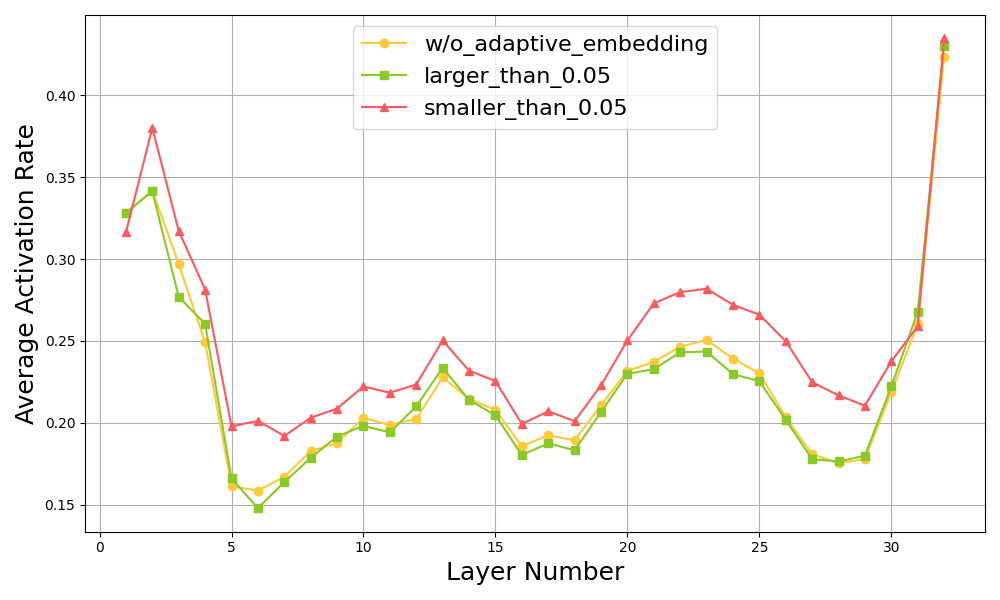}
    \caption{MLP layer activation rates across Transformer layers for the first five tokens of generated answers, sampled 200 times. The curves compare our framework without \embeddingname and with \embeddingname with variance gating larger than 0.05 or smaller than 0.05.}
    \label{fig:activation rate}
\end{figure}

\begin{table}[ht]
    \centering
    \resizebox{\linewidth}{!}{%
    \begin{tabular}{llccc}
        \toprule
        \textbf{Dataset} & \textbf{Method} & \textbf{BM25} & \textbf{DPR} & \textbf{Contriever} \\
        \cmidrule(lr){3-5}
         &  & \multicolumn{3}{c}{Candidates Coverage (\%): $\uparrow$} \\
        \midrule
        \multirow{2}{*}{HotpotQA} 
            & \frameworkname       & \textbf{58.0} & \textbf{50.0} & \textbf{51.0} \\
            & {\bf w/o} exploratory    & 55.0 & 46.0 & 48.2 \\
        \midrule
        \multirow{2}{*}{2Wiki} 
            & \frameworkname       & \textbf{51.0} & \textbf{43.6} & \textbf{31.6} \\
            & {\bf w/o} exploratory    & 47.0 & 40.2 & 28.2 \\
        \midrule
        \multirow{2}{*}{NQ} 
            & \frameworkname       & \textbf{46.6} & \textbf{43.6} & \textbf{45.8} \\
            & {\bf w/o} exploratory   & 43.0 & 41.8 & 41.2 \\
        \midrule
        \multirow{2}{*}{WebQ} 
            & \frameworkname       & \textbf{57.8} & \textbf{56.6} & \textbf{51.8} \\
            & {\bf w/o} exploratory   & 54.6 & 52.0 & 49.2 \\
        \bottomrule
    \end{tabular}%
    }
    \caption{Candidates coverage analysis on four datasets (HotpotQA, 2Wiki, NQ, and WebQ) across three retrievers (BM25, DPR, and Contriever). The metric, \textbf{Candidates coverage (\%)}, represents the proportion of candidates that contain the ground truth answer. We compare our framework \frameworkname with \embeddingname injection against the variant without it (- Exploratory Embedding). The results demonstrate that incorporating \embeddingname injection enhances diversity and increases the likelihood of covering the correct answer.}
    \label{tab:candidate_coverage}
\end{table}

Although we have theoretically demonstrated that minimising the variance allows us to filter out \embeddingname that deviate from the context and query direction (Appendix \ref{appendix:Theoretical Discussion on Embedding Space}), this conclusion may not be entirely intuitive. A natural question arises: why choose 0.05 as the variance gating threshold? Following \citet{naik2024diversity}, we investigate this from the perspective of neuron activations in the Transformer’s MLP layers. Figure \ref{fig:activation rate} illustrates the activation rates across Transformer layers for the first five tokens of generated answers, sampled 200 times. We observe that when \embeddingname is applied with a variance gate lower than 0.05, each layer—particularly from the 5th to the 30th, exhibits an increase in activation rates of roughly 3–4\%. However, when the variance exceeds 0.05, the activation rates remain nearly unchanged or even slightly lower compared to the setting without \embeddingname. This phenomenon suggests that our \embeddingname with the variance gate stochastically triggers more diverse neural pathways.

At the token level, we further explore whether \embeddingname enhances the possibility of generating the correct candidates. Table \ref{tab:candidate_coverage} presents the candidate sentence coverage, defined as the percentage of generated candidate sentences that contain the ground truth answer, across four datasets and three retrievers. We compare our full framework \frameworkname, which incorporates \embeddingname injection, with a variant that excludes this operation ({\bf w/o} Exploratory). The results consistently show that \embeddingname improves coverage across all settings. For example, on HotpotQA with BM25, coverage increases from $55.0\%$ without \embeddingname to $58.0\%$ with it, with similar improvements observed for DPR, Contriever, and across the other datasets. These findings collectively indicate that \embeddingname injection not only promotes more diverse neural activation but also increases the likelihood of including the correct answer in the generated candidates.

\section{Conclusion}
We introduce \frameworkname, an embedding-level framework for open-domain QA that improves efficiency over multi-turn prompt-based systems. By refining query representations with lightweight linear layers trained via unsupervised contrastive learning, our approach reorders retrieved passages to prioritise those most likely to contain correct answers. Additionally, an exploratory embedding with an entropy-based selection mechanism enhances candidate diversity and streamlines self-verification. Experiments across multiple ODQA benchmarks, retrieval methods, and state-of-the-art LLMs show that \frameworkname consistently outperforms prompt-level approaches in accuracy and efficiency.
\section*{Limitations}

Although \frameworkname significantly enhances both efficiency and accuracy in open-domain QA, it comes with several limitations. First, our approach relies on access to an open-source LLM to modify embeddings at the model level, which may not be feasible for scenarios where only black-box API-based models are available. This constraint limits the direct applicability of our method to widely used proprietary models such as GPT-4 or Claude. Second, while \frameworkname reduces the computational overhead associated with multi-turn prompt-based methods, it introduces an additional embedding-level training step. Although this step is lightweight compared to full retriever fine-tuning, it still requires additional computational resources, which may not be ideal for resource-constrained environments. Lastly, our framework assumes that reranking retrieved passages based on learned query refinements will consistently improve answer selection. However, its effectiveness depends on the quality of the retrieved passages—if the initial retrieval fails to include informative passages, reranking alone may not be sufficient to bridge the gap. Future work can explore adaptive retrieval mechanisms to further enhance robustness across diverse retrieval conditions.

\section*{Acknowledgments}
This work was supported in part by the UK Engineering and Physical Sciences Research Council (EPSRC) through a Turing AI Fellowship (grant no. EP/V020579/1, EP/V020579/2) and a New Horizons grant (grant no. EP/X019063/1), and KCL’s Impact Acceleration Account (grant no. EP/X525571/1). A PhD studentship from the Chinese Scholarship Council funds Zhanghao Hu. The authors also acknowledge the use of the King’s Computational Research, Engineering, and Technology Environment (CREATE) at King’s College London. 

\bibliography{anthology,custom}

\clearpage

\appendix

\section{Prompt Design}
\label{sec: appendix prompt design}
In this section, we present the specific prompts used for the experiments in Section \ref{sec:setups}.

\subsection{Answer Candidates Generation}
In Listing 1, we present the prompt $p_{can}$, which is used to generate $K$ answer candidates from the given question and $N$ retrieved passages. Here, we present the case of $K = 2$.
\begin{table}[!htbp]
\centering
\small
% \caption{Prompt for Answer Candidates Generation}
\label{tab:prompt_multi_generate}
\begin{tcolorbox}[width=\columnwidth, title={\textbf{Prompt for Answer Candidates Generation}}, colback=white!95!gray]
\texttt{text = f"Below are \{n\_articles\} passages related to the question at the end. \\ \\
After reading the passages, provide two correct candidates for the answer to the question. \\ \\
Each answer should be in the form: (a) xx, (b) yy, and should not exceed 3 words." \\ \\
Passage \#1 Title: \{Passage \#1 Title\} \\ \\
Passage \#1 Text: \{Passage \#1 Text\} \\ \\
... \\ \\
Passage \#N Title: \{Passage \#N Title\} \\ \\
Passage \#N Text: \{Passage \#N Text\} \\ \\
Question: \{Question\} \\ \\
Answer: \\}
\end{tcolorbox}
\end{table}

\subsection{Prompt Level Rerank Framework Implementation \cite{zhuang-etal-2024-beyond}}
\label{appendix: prompt level rerank framework}

Since \citet{zhuang-etal-2024-beyond} did not release their implementation, we re-implemented their prompt-level rerank framework based on the key ideas outlined in their work. Listing 2 shows the prompt used to rerank passages with LLMs, which includes the query and document text, followed by a relevance judgment instruction on a scale from 0 (Not Relevant) to 4 (Perfectly Relevant), with the output constrained to an integer.
\begin{table}[!htbp]
\centering
\small
\label{tab:prompt_rerank}
\begin{tcolorbox}[width=\columnwidth, title={\textbf{Prompt for Passage Relevance Reranking}}, colback=white!95!gray]
\texttt{Query: \{query\}\\ 
\\
Document: \{doc\_text\}\\
\\
From a scale of 0 to 4, judge the relevance between the query and the document.\\
\\
0 means 'Not Relevant', 1 means 'Little Relevant', 2 means 'Somewhat Relevant', 3 means 'Highly Relevant', 4 means 'Perfectly Relevant'.\\
\\
Return only the integer.
}
\end{tcolorbox}
\end{table}

\section{Theoretical Discussion on Embedding Space}
\label{appendix:Theoretical Discussion on Embedding Space}

For a give set of bounded vectors $\{v_i\}$, the orthogonality can be guaranteed by minimizing the following variable \cite{jain2014almost}:

\begin{equation} \nonumber
    \epsilon = \mathcal{\textbf{max}}_{i\neq j}|v_i\cdot v_j |^2.
\end{equation}

In our work, we suppose the vectors sampled from a LLM are $v$, and the vectors introduced by injecting embeddings to the LLM are $u$. Since there are multiple tokens, including the input tokens and the token will be generated from the LLM, we mark the tokens' vectors as $v_i$. Then, we have an equivalent definition under the LLM setup:

\begin{equation} 
\small
    \epsilon' = \mathcal{\textbf{max}}_{i\neq j,u}\{|v_i\cdot v_j |^2,|v_i,u|^2\} \nonumber.
\end{equation}

\noindent If we have $\forall v_i$, $|v_i,u|^2 \leq \mathcal{\textbf{max}}_{i,j}|v_i\cdot v_j |^2$, obviously we have $\epsilon = \epsilon'$. Otherwise, we obtain a large $\epsilon$ by injecting the embedding. Therefore, we only need to minimize the $\mathcal{\textbf{max}}_{i,u}\{|v_i,u|^2\}$.

According to existing theoretical analysis \cite{geshkovski2024mathematicalperspectivetransformers}, we assume that the $v_i$ is a $k$ dimensional Gaussian vector with mean $\mu_v$ and variance $\sigma_v$ on each dimension. (Empirically, $\mu_v \rightarrow 0$ and $\sigma_v \rightarrow \varepsilon$, where $\varepsilon$ is a small number.)

Then, for any given injected vector $u$, the orthogonality should be decided by $|v_i,u|^2$. However, the $v_i$ contains not only the input tokens' representation but also the prediction of the future. It is computationally expensive to pick a reasonable $u$ after the whole decoding process. It is also meaningless to do so since we have already obtained the decoding results, and no need to discuss whether the injected token embedding is efficient or not. 

So, the problem is, how to estimate the potential inner product with all possible token embeddings for a given vector $u$. Here, we propose to use order statistics to find an equivalent regularisation. 

We first sort the values in each dimension of $u$ into descending order:

\begin{equation} \nonumber
    u' = \{u_{(1)},u_{(2)},u_{(3)},...,u_{(k)}\},
\end{equation}

\noindent where
\begin{equation} \nonumber
    u_{(1)} \geq u_{(2)} \geq u_{(3)} \geq... \geq u_{(k-1)} \geq u_{(k)}.
\end{equation}

Then we duplicate the swapping operation on $v$, which means if we swap the $i$-th dimension with $j$-th dimension on $u$, then the same operation will be deployed on $v$ as well. Here, we suppose there is a general distribution of $v$ on token embeddings, and we mark the duplicating operation results as:

\begin{equation} \nonumber
    v' = \{v_{(1)},v_{(2)},v_{(3)},...,v_{(k)}\},
\end{equation}

Considering the space between two adjacent variables in order statistics $\Delta_i =  u_{(i)} - u_{(i+1)}$ with the only large half where $1\leq i\leq k/2$, we have:

\begin{equation} \nonumber
    \Sigma_{i=1}^{k/2}v_{(i)}u_{(i)} = \Sigma_{i=1}^{k/2}\Sigma_{j=1}^{i}\Delta_iv_{(j)}.
\end{equation}

According to \cite{Boucheron_2012}, we have:

\begin{equation} \nonumber
    \mathbb{E}(\Delta_i^2) \leq \frac{2}{i^2}\mathbb{E}(e^{\frac{u_{(i)}^2}{2}}\int_{u_{(i)}}^{+\infty}e^{-\frac{t^2}{2}}dt).
\end{equation}

Obviously, there are three parts in the expectation of space $\Delta_i$ in the order statistics $u_{(i)}$. With the increasing of order index $i$ in $u_{(i)}$, the value of $u_{(i)}$ is decreasing. Thus, the $2/i^2$ will converge to zero, and the experiential term $e^{u_{i}^2/2}$ will decrease to 1 since most embeddings in LLM are around the original point in current popular models like LLama or Qwen. For the third term of $\int_{u_(i)}^{+\infty}e^{-\frac{t^2}{2}}dt$, it is a survival function from a Gaussian distribution, then it should be bounded by 1. (In fact, it should be bounded by 0.5 if we only consider the positive $u_{(i)}$ from the first half of the order statistics) 

Therefore, for $\forall s>0$, $\exists \varepsilon>0$, when $i>s$, we have $\mathbb{E}(\Delta_i^2) < \varepsilon$ is true. 

Meanwhile, for the whole inner product term, we have 

\begin{small}
\begin{equation} 
\small
|\Sigma_{i=1}^{k}v_{(i)}u_{(i)}|^2 < |\Sigma_{i=1}^{\frac{k}{2}}v_{(i)}u_{(i)}|^2 + |\Sigma_{i=\frac{k}{2}+1}^{k}v_{(i)}u_{(i)}|^2 \nonumber,
\end{equation}
\end{small}

\noindent where we consider both sides of the order statistics, which may contain a large absolute value but negative $u_{i}$.

Then, considering the bound of $\mathbb{E}(\Delta_i^2)$, and $v_{(j)}$ follows a Gaussian distribution with 0 mean for all dimensions, then the expectation of inner product highly relies on the first top $s$ element in the ordered statistics, where:

\begin{equation} \nonumber
    |\Sigma_{i=1}^{k}v_{(i)}u_{(i)}|^2 \leq 4\sigma_v^2\Sigma_{i=1}^s\Delta_i^2\mathbb{E}(e^{\frac{u_{(i)}^2}{2}}) + 2s\sigma_v^2 \varepsilon.
\end{equation}

Therefore, in our proposed method, we design a spacing-based method according to the top $s$ order statistics to approximate the orthogonality of the whole embedding space.

\section{Overconfidence Analysis of Mistral and Qwen}
\label{appendix:Overconfidence Analysis of Mistral and Qwen}

To provide more detail, we break down our framework and analyse where the performance differences in Mistral and Qwen come from.

Our framework consists of two components: a retriever and a reader. As shown in Table \ref{tab:gt_top10} below, all three models achieve similar Top-10 retrieval passages recall for ground-truth passages across the four datasets. This indicates that the limited performance gains for Mistral and Qwen mainly \textit{arise from the reader module}.

\begin{table*}[ht]
\centering
\resizebox{\linewidth}{!}{
\begin{tabular}{l|l|c|c|c|c|c}
\toprule
\textbf{Model} & \textbf{Framework / Avg.GT@Top-10} & \textbf{HotpotQA} & \textbf{2Wiki} & \textbf{NQ} & \textbf{WebQ} & \textbf{Average} \\
\midrule
\multirow{2}{*}{LLaMA3.1-8B-Ins} 
& Retrieval Only w/o Rerank & 1.47 & 0.99 & 1.98 & 2.87 & 1.83 \\
& \textbf{EmbQA (Ours)} & \textbf{1.87} & \textbf{1.49} & \textbf{2.55} & \textbf{4.31} & \textbf{2.56} \\
\midrule
\multirow{2}{*}{Mistral v0.2-7B-Ins} 
& Retrieval Only w/o Rerank & 1.47 & 0.99 & 1.98 & 2.87 & 1.83 \\
& \textbf{EmbQA (Ours)} & \textbf{1.97} & \textbf{1.26} & \textbf{2.90} & \textbf{3.92} & \textbf{2.51} \\
\midrule
\multirow{2}{*}{Qwen 2.5-7B-Ins} 
& Retrieval Only & 1.47 & 0.99 & 1.98 & 2.87 & 1.83 \\
& \textbf{EmbQA (Ours)} & \textbf{2.06} & \textbf{1.39} & \textbf{2.84} & \textbf{4.07} & \textbf{2.59} \\
\bottomrule
\end{tabular}}
\caption{Average ground-truth passages (GT@Top-10) across datasets and models.}
\label{tab:gt_top10}
\end{table*}

\begin{table*}[t]
\centering
\begin{tabular}{l|c|c|c|c|c}
\toprule
\textbf{Model / Avg. Candidate Entropy} & \textbf{HotpotQA} & \textbf{2Wiki} & \textbf{NQ} & \textbf{WebQ} & \textbf{Average} \\
\midrule
LLaMA 3.1 8B Instruct & 1.06 & 0.98 & 0.95 & 0.93 & 0.98 \\
Mistral v0.2 7B Instruct & 0.17 & 0.19 & 0.20 & 0.20 & 0.19 \\
Qwen 2.5 7B Instruct & 0.15 & 0.20 & 0.14 & 0.17 & 0.17 \\
\bottomrule
\end{tabular}
\caption{Average candidate entropy across datasets for different LLMs. Lower entropy indicates higher confidence in selecting relevant candidates.}
\label{tab:candidate_entropy}
\end{table*}

We then examined answer entropy across different models and datasets. On the four context-based QA datasets used in our experiments, both Mistral and Qwen tend to produce answers with very high confidence scores (i.e., predictive probabilities). This observation is supported by the results in Table \ref{tab:candidate_entropy}, where we measure the entropy of candidate phrases across datasets. Mistral and Qwen consistently yield \textit{lower entropy values}, indicating a strong tendency toward \textit{overconfident predictions}.

We observe that this overconfidence causes Mistral and Qwen to produce a highly skewed distribution when fed with the proposed exploratory embedding. As a result, the sum of differences we compute is significantly smaller than the theoretical estimation. This suggests that the theoretical analysis-based unified threshold used in the experimental setup for all models—based on the classical 5\% significance level—may not be effective for these two models. This explains why the improvements for Mistral and Qwen were not statistically significant. Nevertheless, a key advantage of this theoretically derived threshold is that it reliably provides a performance lower bound across different models, ensuring robustness even when improvements are not statistically significant.

Moreover, we explored performance adjustments by selecting appropriate hyperparameter values for threshold T. To explore this further, we conducted ablation studies on the HotpotQA dataset, serving as a development set, with varying threshold values (Table \ref{tab:threshold_em_f1}) and the Contriever-based retriever. Results show that different models favour different thresholds: Mistral performs best at 0.01, while Qwen performs best at 0.005. This indicates that a fixed threshold may not be optimal across models.

\begin{table}[t]
\centering
\resizebox{\linewidth}{!}{
\begin{tabular}{l|c|c|c|c}
\toprule
\textbf{Model / Threshold} & \textbf{0.05} & \textbf{0.01} & \textbf{0.005} & \textbf{0.001} \\
\midrule
Mistral v0.2 7B Ins & 29.8 / 42.3 & \textbf{31.2 / 43.6} & 30.6 / 42.7 & 30.3 / 43.5 \\
Qwen 2.5 7B Ins     & 39.0 / 50.2 & 39.8 / 51.5 & \textbf{42.0 / 52.8} & 40.8 / 51.9 \\
\bottomrule
\end{tabular}
}
\caption{Effect of varying the selection threshold on EM/F1 performance for Mistral and Qwen on HotpotQA dataset.}
\label{tab:threshold_em_f1}
\end{table}

\section{Further Fusion Study on Contrastive Learning and the Embedding Refinement}

\label{appendix:Further Fusion Study on Contrastive Learning and the Embedding Refinement}

To clarify whether performance gains originate primarily from contrastive learning or the embedding refinement itself, we conducted an additional ablation experiment (The ``New experiment" of the table \ref{tab:ablation_embqa_llama}). Specifically, we removed both the exploratory embedding and contrastive learning modules from our reranking framework, directly summing the query and candidate representations for reranking. As shown in the table, this approach yields minimal improvement, highlighting that the significant gains indeed arise from \textit{integrating contrastive learning with embedding refinement}.

\begin{table*}[t]
\centering
\resizebox{\linewidth}{!}{
\begin{tabular}{l|l|c|c|c|c|c}
\toprule
\textbf{LLaMA 3.1 8B Instruct} & \textbf{EM / F1} & \textbf{HotpotQA} & \textbf{2Wiki} & \textbf{NQ} & \textbf{WebQ} & \textbf{Average} \\
\midrule
From Figure 2 & EmbQA w/o exploratory \& reranking & 33.6 / 50.2 & 22.2 / 31.1 & 39.6 / 53.7 & 34.4 / 48.3 & 32.5 / 45.8 \\
\textbf{New Experiment} & \makecell[l]{EmbQA w/o exploratory \& \\ contrastive learning in rerank} & 33.8 / 50.5 & 22.4 / 31.4 & 40.2 / 53.0 & 34.0 / 49.1 & 32.6 / 46.0 \\
From Figure 2 & EmbQA w/o exploratory embedding & 35.2 / 50.9 & 23.0 / 31.7 & 41.8 / 53.5 & 34.8 / 49.3 & 33.7 / 46.4 \\
From Figure 2 & \textbf{EmbQA (Ours)} & \textbf{36.6 / 52.7} & \textbf{26.4 / 34.2} & \textbf{42.2 / 53.6} & \textbf{36.0 / 49.6} & \textbf{35.3 / 47.5} \\
\bottomrule
\end{tabular}
}
\caption{Ablation results (EM / F1) of EmbQA under different component removal settings on LLaMA 3.1 8B Instruct.}
\label{tab:ablation_embqa_llama}
\end{table*}

\section{Detailed Comparison of Multi-turn Prompting Costs}

\label{appendix:Detailed Comparison of Multi-turn Prompting Costs}

To clarify the computational advantages of EmbQA, we provide a step-by-step breakdown of the prompting structure and cost comparison with SuRe~\cite{kim2024sure}, as summarised below.

\paragraph{SuRe (7 prompts per question):}
\begin{itemize}
    \item \textbf{Step 1: Candidate Generation (1 prompt)} — Generate $K=2$ initial answers from retrieved passages.
    \item \textbf{Step 2: Summary Generation (2 prompts)} — Generate a summary for each candidate.
    \item \textbf{Step 3: Validity Measurement (2 prompts)} — Assess factuality of each summary individually.
    \item \textbf{Step 4: Ranking (2 prompts)} — Conduct pairwise comparisons of summaries in both orders to avoid bias.
\end{itemize}

\paragraph{EmbQA (2 prompts per question):}
\begin{itemize}
    \item \textbf{Step 1: Candidate Generation (1 prompt)} — Generate two initial answers from retrieved passages.
    \item \textbf{Step 2: Re-ranking (0 prompt)} — Contrastive re-ranking is performed without LLM prompting.
    \item \textbf{Step 3: Exploratory Embedding (1 prompt)} — Inject exploration signals and generate diverse candidates.
    \item \textbf{Step 4: Entropy-based Selection (0 prompt)} — Final answer selected based on entropy, without prompting.
\end{itemize}

\paragraph{Cost Scaling with Candidate Size.}
SuRe scales linearly in prompt cost with the number of candidates, requiring 3 additional prompts per candidate beyond $K=2$. In contrast, EmbQA maintains a constant prompt count regardless of candidate number due to its reranking and selection mechanisms being handled without LLM queries.

\paragraph{Summary.}
EmbQA achieves comparable or better performance using only 2 prompts per question, while SuRe requires 7. Considering both prompt count and per-step complexity, EmbQA operates at roughly one-third the runtime and output token cost of SuRe on average.

Based on the above steps, we report a detailed statistical comparison in the HotpotQA dataset as an example. Time/query means execution time per query (lower is better), and Token/query means output token requirement per query (lower is better). 

\begin{table}[t]
\centering
\resizebox{\linewidth}{!}{
\begin{tabular}{l|l|c|c}
\toprule
\textbf{Method} & \textbf{Step} & \textbf{Time/query (min)} & \textbf{Token/query} \\
\midrule
\multirow{5}{*}{SuRe} 
  & Step 1 & 0.22 & 0.49k \\
  & Step 2 & 0.45 & 1.01k \\
  & Step 3 & 0.47 & 1.06k \\
  & Step 4 & 0.42 & 0.95k \\
  & \textbf{Total} & \textbf{1.56} & \textbf{3.51k} \\
\midrule
\multirow{5}{*}{EmbQA (Ours)} 
  & Step 1 & 0.23 & 0.50k \\
  & Step 2 & 0.08 & 0 \\
  & Step 3 & 0.21 & 0.49k \\
  & Step 4 & 0.01 & 0 \\
  & \textbf{Total} & \textbf{0.53} & \textbf{0.99k} \\
\bottomrule
\end{tabular}
}
\caption{Step-wise comparison of execution time and output token cost per query between SuRe and EmbQA on HotpotQA.}
\label{tab:efficiency_breakdown}
\end{table}

\section{Reranking with Closed-Source API Models}
\label{appendix:closedsource_rerank}

Although our full EmbQA pipeline requires access to model internals (e.g., exploratory embedding and entropy-based selection), the reranking module alone can be integrated with closed-source API models. To validate this, we conduct experiments using the GPT-4o-mini model with Contriever retrieval. We compare three configurations: retrieval-only, SuRe~\cite{kim2024sure}, and SuRe enhanced with our embedding-level reranker. Results across four ODQA datasets (Table~\ref{tab:gpt4o_rerank}) demonstrate consistent improvements in EM and F1, highlighting the generalisation potential of our reranking design.

\begin{table}[h]
\centering
\resizebox{\linewidth}{!}{
\begin{tabular}{l|c|c|c|c|c}
\toprule
\textbf{Framework} & \textbf{HotpotQA} & \textbf{2Wiki} & \textbf{NQ} & \textbf{WebQ} & \textbf{Average} \\
\midrule
Retrieval-only & 43.8 / 60.6 & 31.0 / 41.2 & 43.0 / 58.1 & 33.4 / 50.6 & 37.8 / 52.6 \\
SuRe & 44.8 / 59.7 & 34.4 / 45.1 & 48.4 / 60.5 & 38.8 / 52.6 & 41.6 / 54.5 \\
\textbf{\makecell[l]{SuRe + \\ EmbQA rerank}} & \textbf{46.9 / 62.5} & \textbf{35.7 / 47.5} & \textbf{50.9 / 62.7} & \textbf{42.1 / 54.8} & \textbf{43.9 / 56.9} \\
\bottomrule
\end{tabular}
}
\caption{Effectiveness of the proposed embedding-level rerank module on GPT-4o-mini across four ODQA datasets.}
\label{tab:gpt4o_rerank}
\end{table}

\end{document}